\documentclass[conference]{IEEEtran}
\IEEEoverridecommandlockouts
\usepackage{cite}
\usepackage{amsmath,amssymb,amsfonts}
\usepackage{algorithmic}
\usepackage{graphicx}
\usepackage{textcomp}
\usepackage{xcolor}
\usepackage{hyperref}
\usepackage{multirow}

\def\BibTeX{{\rm B\kern-.05em{\sc i\kern-.025em b}\kern-.08em
    T\kern-.1667em\lower.7ex\hbox{E}\kern-.125emX}}
\begin{document}

\title{Where Is My Mind (looking at)? Predicting Visual Attention from Brain Activity}

\author{

\IEEEauthorblockN{Victor Delvigne}
\IEEEauthorblockA{\textit{ISIA Lab, Faculty of Engineering} \\
\textit{University of Mons}\\
Mons, Belgium \\
\IEEEauthorblockA{\textit{IMT Lille Douai} \\
\textit{CRIStAL UMR CNRS 9189}\\
Villeneuve d’Ascq, France\\
victor.delvigne@umons.ac.be}\\
}

\and

\IEEEauthorblockN{Noé Tits}
\IEEEauthorblockA{\textit{ISIA Lab} \\
\textit{ Faculty of Engineering} \\
\textit{University of Mons}\\
Mons, Belgium}\\

\and

\IEEEauthorblockN{Luca La Fisca}
\IEEEauthorblockA{\textit{ISIA Lab} \\
\textit{ Faculty of Engineering} \\
\textit{University of Mons}\\
Mons, Belgium}\\

\and

\IEEEauthorblockN{Nathan Hubens}
\IEEEauthorblockA{\textit{ISIA Lab} \\
\textit{ Faculty of Engineering} \\
\textit{University of Mons}\\
Mons, Belgium}\\

\and

\IEEEauthorblockN{Antoine Maiorca}
\IEEEauthorblockA{\textit{ISIA Lab} \\
\textit{ Faculty of Engineering} \\
\textit{University of Mons}\\
Mons, Belgium}\\

\and

\IEEEauthorblockN{Hazem Wannous}
\IEEEauthorblockA{\textit{IMT Lille Douai} \\
\textit{CRIStAL UMR CNRS 9189}\\
Villeneuve d’Ascq, France}\\

\and

\IEEEauthorblockN{Thierry Dutoit}
\IEEEauthorblockA{\textit{ISIA Lab} \\
\textit{ Faculty of Engineering} \\
\textit{University of Mons}\\
Mons, Belgium}\\

\and 

\IEEEauthorblockN{Jean-Philippe Vandeborre}
\IEEEauthorblockA{\textit{IMT Lille Douai} \\
\textit{CRIStAL UMR CNRS 9189}\\
Villeneuve d’Ascq, France}\\

}
\maketitle

\begin{abstract}
Visual attention estimation is an active field of research at the crossroads of different disciplines: computer vision, artificial intelligence and medicine. One of the most common approaches to estimate a saliency map representing attention is based on the observed images. In this paper, we show that visual attention can be retrieved from EEG acquisition. The results are comparable to traditional predictions from observed images, which is of great interest. For this purpose, a set of signals has been recorded and different models have been developed to study the relationship between visual attention and brain activity. The results are encouraging and comparable with other approaches estimating attention with other modalities. The codes and dataset considered in this paper have been made available at \url{https://figshare.com/s/3e353bd1c621962888ad} to promote research in the field.
\end{abstract}

\begin{IEEEkeywords}
component, formatting, style, styling, insert
\end{IEEEkeywords}

\section{Introduction}

Saliency heatmap estimation is a field of research at cutting edge of technology today. Estimating with precision the region of the field of view where human is focused is a great help for many computer vision applications. In most of the works aiming to estimate images that represents the region of interest in the field of view, also called visual saliency map, the considered modalities are often images and videos \cite{droste_unified_2020,pan_salgan_2018}.

Nowadays, machine learning (ML) and the topics deriving from it have known a huge increase in interest. More and more publications and research projects related to novel deep learning (DL) algorithms in the context of computer vision or natural language processing (NLP) have been presented in recent years. Although ML algorithms tend to be used for those fields, a growing interest has been noticed in the medical domain \cite{ravi_deep_2017}. Moreover, the use of ML algorithms may be an interesting opportunity to improve diagnosis, help the works of specialists and to have a better understanding of biomedical signals.

As of today, the existing works aiming to estimate visual saliency are based on images. It could be interesting to exploit the scientific research proving the relationship between brain activity and attention mechanism \cite{duncan_competitive_1997} by estimating visual saliency from biomedical signals. The goal being not to beat the results provided by image-based methods but to investigate this novel relationship.

On another hand, the increasing amount of data and their democratisation have to led to an increase of research projects in Brain-Computer Interfaces (BCI) which aims to promote interaction between the human brain and the computer. This connection can be (non-)invasive and more or less expensive depending on the considered biomedical signals. Among the different types of biomedical signals considered in existing research projects, electroencephalogram (EEG) representing electrical brain activity, seems to be prone for this type of applications. The motivations being based on their ease of use, relatively low cost compared to other techniques while maintaining a high fidelity for signals acquisition.

In this context, it has been considered to investigate the relationship between eye-tracking and EEG. For this purpose, we propose a novel framework aiming to estimate visual saliency map from electrophysiological recordings. The contribution of this paper can be summarized in 3 points: (1) an adaptation for raw signals of the existing methods for EEG's features representation under images form; (2) a novel feature extraction method representing EEG signals in lower subspace; (3) a framework estimating visual saliency map from electrophysiological signals. 


\section{Related Work}
\label{sec:RW}
The related work has been split into three subsections: (1) Deep learning approaches for EEG processing, presenting different methods based on DL to process EEG; (2) EEG-based attention estimation, introducing the research projects related to attention in the context of EEG; (3) Visual saliency estimation, showing the existent works aiming to estimate saliency map from several modalities.

\subsection{Deep learning approaches for EEG processing}
As previously mentioned, ML algorithms have known an increase in interest for some time now. It has also been the case in the context of biomedical signal processing and brain imaging researches. More specifically, in the case of electroencephalogram signals processing, several deep learning approaches have been considered for different purposes \cite{lotte_review_2018}. In most of the cases, EEGs are considered as an array $ X \in  \mathbb{R}^{t \times elec}$  with $t$ representing the time evolution and $elec$ the number of considered electrodes. A non-exhaustive list of works considering DL algorithms with EEG is the following:
\begin{itemize}
    \item The use of convolutional neural networks (CNN) has been considered to extract feature from EEG signals. One of the best known models is EEGNet presented by Lawhern et al. \cite{lawhern_eegnet_2018}. This network aims to estimate motor movements and detect evoked potentials (specific pattern in electrophysiological signals seen after stimuli apparition) through a sequence of convolution filters with learnable kernels. These kernels extract the spatial and/or temporal features from the signal according to the considered shape (x-axis representing the time evolution and y-axis the considered channels). 
    
    \item One of the other methods considered to process EEG is the use of graph networks. With this approach, the EEG is considered as a graph (with vertices corresponding to electrodes and edges being proportional to their distance) evolving over time. The method based on Regularized Graph Neural Networks proposed by Zhong et al. \cite{zhong_eeg-based_2020} presents the best results for emotion estimation from EEG. 
    
    \item Another approach that has already been considered for a wide range of application in EEG processing is based on recurrent neural networks (RNN). These kinds of networks have already proven their ability to extract the spatial \cite{li_novel_2020} and temporal \cite{bashivan_learning_2015} information from brain activity signals. In the work of Bashivan et al. \cite{bashivan_learning_2015}, they consider a model composed of a different layer of CNN and RNN to estimate motor movements from EEG. 
    
    \item Over the last years, an emerging method have been considered: the use of a generative adversarial network (GAN) for EEG processing. GANs have already been used for generating images representing thougths and/or dreams \cite{tirupattur_thoughtviz_2018,palazzo_generative_2017}. Although this research field is still under development, the authors have high hopes that one day, it will be possible to visualise our thoughts or dreams.
\end{itemize}

In addition to the different DL models used for estimation and regression from brain activity, it is also possible to consider different feature extraction and representation methods. In \cite{lawhern_eegnet_2018}, they directly considered the raw signal and let the models extracting the most significant feature. On the other hand, in \cite{bashivan_learning_2015,li_novel_2020}, they considered well-known feature extraction methods expressing the spectral \cite{bashivan_learning_2015} and temporal information \cite{li_novel_2020,zhong_eeg-based_2020} from signals. Moreover, in \cite{bashivan_learning_2015} they consider a more visual representation of EEG feature under a more understandable form. In their approach, they consider the position of each electrode in the 3D frame and create an image within which the location of pixels and electrodes are correlated and their value is related to the feature value in the specific location.

\subsection{EEG-based attention estimation}
Liang et al. \cite{liang_characterization_2018} present an approach to estimate visual saliency features from EEG. The considered methodology consists of a joint recording of EEG while watching video clips. Saliency features representing the degree of attention and average position of the centre of interest in the video. The presented results were encouraging for further study and indicate the existence of a relationship between visual attention and brain activity. On the other hand, different datasets aiming to estimate the attention state from biomedical signals have been published. Cao et al. \cite{cao_multi-channel_2019} and Zheng et al. \cite{zheng_multimodal_2017} considered recordings of EEG and eye-tracking signals to estimate the attention state of a participant during specific tasks. Zheng et al. \cite{zheng_multimodal_2017} show that it is possible to estimate the attention state from these joint recordings in many cases. 

The lack of in-depth studies aiming to investigate the relationship between EEG signals and visual saliency has been a motivation for the creation of a framework aiming to investigate the relationship between these two modalities.

\subsection{Visual saliency estimation}
Visual saliency estimation is a field at cutting edge in computer vision domain. There exist a lot of different models aiming to estimate visual saliency from different modalities. In most of the cases, the goal of these models is to estimate the visual saliency region from images as reported in the MIT/Tuebingen Saliency Benchmark \cite{kummerer_saliency_2018} including all the existing models. Among the existing works, a certain amount of the proposed methods was based on the succession of an encoding (composed of successions of convolution and max-pooling layers) and decoding (resp. succession of convolution layer and upsampling layers) networks as in the works of Kroner et al. \cite{kroner_contextual_2020} and Pan et al. \cite{pan_salgan_2018} being in the best results among the SOTA works.

\begin{figure}[t]
    \begin{center}
     \includegraphics[width=1\linewidth]{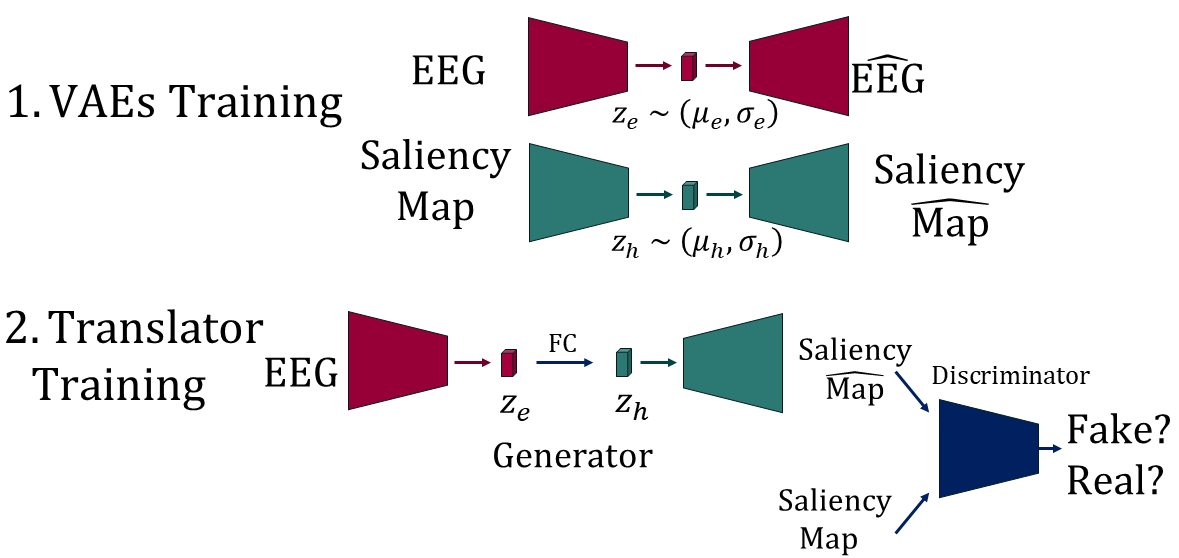}
    \end{center}
       \caption{Overview of the proposed framework with the three models considered: 1) VAE for EEG signals; 2) VAE for saliency map; 3) GAN estimator.}
    \label{fig:framework}
\end{figure}

\section{Proposed Method}

The goal of our work is to combine the existing DL methods to study the relationship between electrophysiological signals and visual saliency map. The framework is divided into three models: 
\begin{itemize}
    \item a variational autoencoder (VAE) aiming to represent the saliency images in a shorter subspace called latent-space. This VAE will have two roles: re-creating the saliency images that represents the participant visual attention; representing the images in a completed and continuous latent-space \cite{kingma_auto-encoding_2014}. 
    \item a VAE aiming to represent the EEG in the latent-space. As for the previous model, the aim of EEG VAE is also to minimize the error between the EEG and its reconstruction and to create a continuous and completed representation of the signals in the corresponding latent-space. 
    \item a GAN binding the EEG and map latent-space with the help of the two VAEs already describe. Besides, a discriminator is also used to classify the images from synthetic (i.e. created by our model) vs. real (i.e. real saliency maps created from eye-tracking recording). 
\end{itemize}
Fig. \ref{fig:framework} shows the framework pipeline separated in two steps: the training of both VAEs and the training of the saliency generator part. The choice of considering VAE instead of the conventional deep autoencoder is motivated by the fact that there is no one-to-one relationship between brain activity and visual attention (by one-to-one relationship Stephani et al.\cite{stephani_temporal_2020} mean that one and only one brain activity corresponds to one and only one map representing attention). This phenomenon, showing that different brain activation may correspond to a single task (and vice-versa), has been studying by Stephani et al.\cite{stephani_temporal_2020}. Moreover, this aspect is enhanced by the fact that it is difficult to extract information from electrophysiological signals due to their trend to be prone to noise and artefacts. The use of VAE instead of conventional AE enables to estimate the distribution (characterised by mean and standard deviation) of the latent space and to study the relationship between their distribution instead of creating a one-to-one relationship between latent vector. 

We separated the proposed methods into four subsections each of them being a specific step of our work: (1) Autoencoding Saliency Map; (2) Autoencoding EEG Signals; (3) Translation Network mapping the latent-spaces distributions; (4) Training Methodology.

\subsection{Autoencoding saliency map}
From raw eye-tracker recordings, it is possible to create a visual saliency map representing the area of attention in an image of one channel with values between 0 and 1 representing the degree of visual attention on specific pixels and their neighbours. It can also be considered as a probability for a given pixel to be watched or not.

During the experimentation, eye-tracker has been jointly recorded with EEG. First, the recordings have been separated into trials corresponding to a specific time. Then, the discrete eye-tracking measurements have been projected on 2D images (one per trials). After, the eye-tracker accuracy has been taken into account by considering circles of radius proportional to the error rate despite discrete points. Finally, Gaussian filtering has been applied to the images with a kernel size ratio corresponding to the eye-tracker field of view. The images generation have been inspired by Salvucci et al. \cite{salvucci_identifying_2000}.

Given the visual saliency images, a VAE has been trained to represent them in a lower subspace. The considered network architecture is based on the ResNet proposed by He et al. \cite{he_deep_2015}. We consider for the encoding part four stacks of ResNet layer, each composed of three convolutional layers and batch normalization, each stack except the last being separated by a max-pooling operation. A similar approach has been considered for the decoding part with upsampling layer instead of a max-pooling operation, the padding has been adapted to ensure that the output size match the input. 

The goal of this network is double: (1) recreating an image as faithful as possible to the original saliency map via a representation in shorter latent space; (2) creating a continuous and complete latent-space, and therefore not to favour one dimension among others. 

After some experimental tests, it has been constated that the VAE tend to slightly overfit after a certain amount of epochs. To reduce this issue and to build a more robust network, a data augmentation policy has been considered. To keep the physical behaviour behind visual saliency map, the data augmentation process had to be well designed. For this reason, we have considered for each training sample of each batch from the training set a random horizontal flip with a probability of 0.5 (the stimuli being equally disposed at the right and left part of the screen as shown in Subsection \ref{ssection:SigAcq}) and a random vertical and horizontal translation between -5 and +5 pixels. This method helped to generate a wider range of visual saliency map with a lower error rate between the initial image and its reconstruction with a better representation of the latent space.

\begin{figure*}
    \begin{center}
    \includegraphics[width=1.05\linewidth]{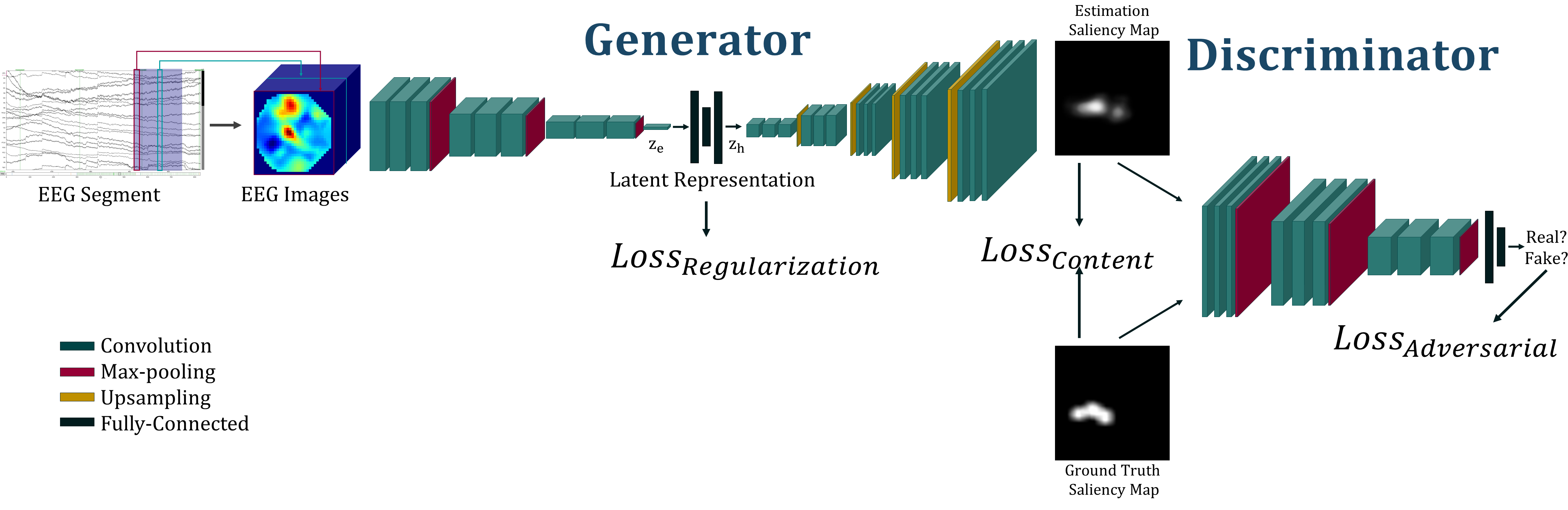}
    \end{center}
   \caption{Overview of the architecture composed of the generator + discriminator model. As seen in the figure, the different key steps are represented from left to the right: the creation of EEG-Images, encoding of EEG-Images, Distribution mapping to saliency map latent-space, decoding to estimate the saliency map and discrimination between generated and ground truth saliency map.}
\label{fig:Network}
\end{figure*}

\subsection{Autoencoding EEG signals}
\label{ssection:Sub_VAE_EEG}
EEG can be considered as two-dimension time series, the first corresponding to time evolution and the second to the considered electrodes. A similar approach than the one proposed by \cite{bashivan_learning_2015} has been considered. This last consists of the creation of images representing the features from EEG according to their spatial location on the participant scalp. The process to construct the images is the following: 
\begin{itemize}
    \item Separating the total recording in trials leading to an array of dimension $[n_{trials} \times t \times n_{electrodes}]$.
    \item For each trial, downsampling the signals after low-pass filtering to extract the general signal evolution and to ignore the artefacts contribution. A preprocessing is at the same time applied to remove the noise and remaining artefacts. 
    \item From the regular electrode position on the scalp, an azimuthal projection is applied to represent their location in a 2D frame. 
    \item The samples composing the trials are taken separately. These samples are projected in a 2D coordinate frame as mentioned at the previous step and a bicubic interpolation is applied to consider a continuous representation of information. The process is repeated for all the samples and each projection are concatenated to lead to an image with a number of channels corresponding to the number of temporal samples after the signal downsampling. 
    \item The statistical distribution of the images set has also been normalized around a mean of 0 and a standard deviation of 1. 
\end{itemize}
This maps representation of EEG allows keeping the spatial (in the first and second dimension) and temporal (between channels) relationship between samples. For clarirty, the maps generated from EEG are mentionned as EEG images in the paper. Moreover, this methodology better suits for CNN than the array representation of EEG. It enables the consideration of squared shape kernels unlike the older models considering unidimensional kernels for feature extraction from EEG \cite{lawhern_eegnet_2018}. 

In a similar way to the image VAE, the EEG images have been passed through a VAE to reduce the EEG dimension and to represent them in a continuous and completed subspace. For this purpose, the VAE has been trained with the images-based EEG. 

A similar methodology, than for the saliency map, has been considered to construct the most robust network as possible. To that end, a random signal following a gaussian distribution of zero mean and standard deviation = 0.25 has been added to EEG images to increase the model stability and to have a better understanding of the difference between noise and EEG. In addition, some pixels composing the EEG images have been supposed to remain equal to zero, these groups of pixels corresponding to the region of the space where there are no electrodes. A checking was set up to verify that those regions of the images remained equal to zero.

\subsection{Generator network mapping the latent-space distributions}
From the representation of the saliency map and EEG in their corresponding shorter subspace, the possibility of mapping the two distributions has been investigated. For this purpose, several approaches have been tested, however, for the paper clarity, only the ones presenting the best results have been presented. 

As explained above, one of the most important barriers in this paper was to create a model permitting the estimation of a saliency map from EEG without considering a one-to-one correspondence between modalities. To solve this issue, a GAN approach has been considered with a generator aiming to recreate the image latent representation from EEG latent representation combined with a discriminator aiming to distinguish the images generated by the generator (composed of the combination of the encoding and decoding part of the VAEs presented in the previous subsection). Also, as it is the case in several approaches \cite{goodfellow_nips_2017}, noise following a normal centred distribution (i.e. mean = 0 and std = 1) has been concatenated to the latent vector at the centre of the generator. This concatenation aims to guide the generator for the saliency map generation. 

The overall architecture of the networks aiming to translate EEG space into image space is represented in Fig. \ref{fig:Network}. As seen in this figure, the generator consists of the concatenation of the encoding part of EEG VAE and decoding part of Saliency VAE through a generator composed of fully-connected (FC) layers. Moreover, as mentioned above a discriminator has also been placed at the end of the model.

\subsection{Training methodology}
As already mentioned, the generator model consists of the concatenation of two parts of pre-trained models. The training policy can be considered in several key steps: 

The training separately of the two VAE. During this training, the goal is to reduce a combination of a content loss, aiming to reduce the reconstruction error, and a regularization loss, promoting a continuous and completed distribution in the latent subspaces. The considered content loss for Image VAE is the binary cross-entropy, the values of the images being included between 0 and 1. For the EEG VAE, a MSE loss has been considered for the opposite reason (EEG images taking values $<0$ and $>1$). The minimisation policy for the two models can be formulated as for the heatmap: 
    \begin{equation}
        \begin{aligned}
            \mathcal{L}_{h} (\theta_h)= \mathcal{L}_{cont} + \mathcal{L}_{reg} \\
            = BCE(y_{t}, y_{p}) + 0.5*KLD(\mu_i, log(\sigma_i))
        \end{aligned}
    \end{equation}
    with $\theta_h$ the model parameters of the image VAE, $y_{t}$ the ground truth saliency map, $y_{p}$ the reconstructed map, $\mu_i$ and $\sigma_i$ being the mean and variance of the estimated latent-space. For the losses, BCE being the binary cross entropy loss and KLD the Kullback-Leibler divergence \cite{kingma_auto-encoding_2014}. And for the EEG:
    \begin{equation}
        \begin{aligned}
            \mathcal{L}_{e} (\theta_e)= \mathcal{L}_{cont} + \mathcal{L}_{reg} \\
            = MSE(y_{t}, y_{p}) + 0.5*KLD(\mu_i, log(\sigma_i))
        \end{aligned}
    \end{equation}
with $\theta_h$ the model parameters of the image VAE, $y_{t}$ the ground truth EEG, $y_{p}$ the reconstructed EEG, $\mu_i$ and $\sigma_i$ being the mean and variance of the estimated latent-space.
During the training of both VAE, a phase of sampling is necessary to estimate the current latent projection from its mean and standard deviation. For this purpose, we use the reparameterization trick \cite{kingma_auto-encoding_2014}, this last considering a random variable following a Gaussian distribution with mean = 0 and standard deviation = 1, to solve the backpropagation issue that the direct sampling cause. 

The generator model has then been created by concatenating the two parts of the VAE and linking them with FC layers. Moreover, the discriminator has also been added after the decoding part. After the architecture creation, all the weights composing the decoder and encoder parts have been fixed (i.e. no gradient descent has been applied on those parameters).  An exception has been made for the last layers of the encoder and the first layers of the decoder (i.e. model parts directly connected to the latent-space) to promote the networks fine-tuning. 

During the training of the generator, three losses have been considered: a content loss to assess the similarity between the ground truth and the created maps; an adversarial loss aiming to distinguish the generated vs. the ground truth; a regularization loss assessing that the latent-space representation remains continuous and complete. The competition between the generator and the discriminator helps for the generation of maps being difficult to differentiate with the ground truth. The considered metrics for each of these losses were: the BCE for the content loss and the adversarial loss and KLD for the regularization loss. The minimisation paradigm to train the generator can be formulated as: 
\begin{equation}
    \begin{aligned}
        \mathcal{L} (\theta)= \mathcal{L}_{cont} + \mathcal{L}_{reg} + \mathcal{L}_{adv} \\
        = BCE(y_{t}, y_{p}) + 0.5*KLD(\mu_i, log(\sigma_i)) + \\ BCE(D(y_{t}), 1) + BCE(D(y_{p}), 1)
    \end{aligned}
\end{equation}
with $\theta_h$ the model parameters, $y_{t}$ the ground truth saliency map, $y_{p}$ the reconstructed map, $\mu_i$ and $\sigma_i$ being the mean and variance of the estimated latent-spaces and $D(_{})$ being the output of the discriminator.

\section{Experiments}
In this section, we first present the considered methodology for the joint acquisition of EEG and eye-tracking signals. Then, the implementation details of the models will be described and finally, the proposed metrics for our approach evaluation will be presented in the last subsection.

\subsection{Dataset acquisition}
\label{ssection:SigAcq}
The signals considered in this paper have been acquired on 32 healthy participants during 15 minutes long session, each participant has made only one session. The signal acquisition protocol consists of a short video game in virtual reality (VR) where it is asked to the participant to direct his gaze on stimuli corresponding to object appearing in a random position of the field of view. 

During the proceeding of the tasks, the eye position and EEG has been synchronously registered. EEG has been recorded with a 32 electrodes biomedical headset (actiCHamp from BrainVision) following the 10/20 electrodes disposition \cite{oostenveld_five_2001}. On another hand, the eye tracker has been registered with the dedicated device placed into the VR headset (HTC Vive Pro Eye with an integrated eye-tracker). After the registration, trials have been extracted from the total recording by segmenting both signals around the stimuli apparition: we consider for both physiological signals the beginning of the segment one second before and the end three seconds after the stimuli appearance. 

This dataset has been record in the context of research project aiming to investigate the relationship between attention and brain activity. For this purpose, a pipeline aiming to record physiological signals in VR has been designed. To promote results reproducibility and research in this domain, the signals acquired has been made publicly available.

\begin{table}
    \begin{center}
        \begin{tabular}{l c c}
            {} & \textbf{Generator}& {} \\
            \hline
            \hline
            Network Part & Layer & Int/Out Channels\\
            \hline
            \multirow{14}{*}{EEG Encoder}   & Conv2D-3 & 401/256\\
                                            & Conv2D-3 & 256/256\\
                                            & Conv2D-3 & 256/256\\
                                            & Conv2D-3 & 256/256\\
                                            & Maxpool-2 & 256/256\\
                                            & Conv2D-3 & 256/512\\
                                            & Conv2D-3 & 512/512\\
                                            & Conv2D-3 & 512/512\\
                                            & Maxpool-2 & 512/512\\
                                            & Conv2D-3 & 512/512\\
                                            & Conv2D-3 & 512/512\\
                                            & Flattening Layer & {}\\
                                            & Linear & 512/256\\
                                            & Linear & 256/64\\
            \hline 
            \multirow{5}{*}{Distrib. FC}   & Linear & 64/64\\
                                            & Sampling Layer & {} \\
                                            & Noise Cat & 64/128 \\
                                            & Linear & 128/256 \\
                                            & Linear & 256/64 \\
            \hline 
            \multirow{20}{*}{Saliency Decoder}  & Linear & 64/512\\
                                                & UnFlattening Layer & {} \\
                                                & Upsampling-2 & 512/512\\
                                                & Conv2D-3 & 512/512\\
                                                & Conv2D-3 & 512/512\\
                                                & Conv2D-3 & 512/512\\
                                                & Upsampling-3 & 512/512\\
                                                & Conv2D-3 & 512/256\\
                                                & Conv2D-3 & 256/256\\
                                                & Conv2D-3 & 256/256\\
                                                & Upsampling-3 & 256/256\\
                                                & Conv2D-3 & 256/128\\
                                                & Conv2D-3 & 128/128\\
                                                & Conv2D-3 & 128/128\\
                                                & Upsampling-3 & 128/128\\
                                                & Conv2D-3 & 128/64\\
                                                & Conv2D-3 & 64/64\\
                                                & Conv2D-3 & 64/64\\
                                                & Upsampling-2 & 64/64\\
                                                & Conv2D-3 & 64/4\\
                                                & Conv2D-3 & 4/4\\
                                                & Conv2D-3 & 4/1\\
            \hline
        \end{tabular}
    \end{center}
\caption{Architecture of the generator network.}
\label{tab:ArchGen}
\end{table}

\subsection{Implementation details}

Given the raw EEG considered as an array of dimension $[n_{trials} \times n_{channels} \times n_{samples}]$ with $n_{trials} = 4000$ being the total number of trials for all the sessions, $n_{channels} = 32$ being the total number of electrodes of the EEG cap and $n_{sample} = t_{length} \times f_{sampling} = 4 * 500  = 2000$ being the total number of samples composing the EEG signal. The signals have been down-sampled with a ratio $=5$, and passed through a low-pass filter with cut-off frequency $= 35 Hz$. Then EEG-images have been created from the preprocessed signals with the methodology explained in section \ref{ssection:Sub_VAE_EEG}, the result dimension being $[n_{trials} \times n_{down} \times h \times w]$ with $n_{down} = 401$ being the amount of samples after down-sampling and $h=32$ and $w=32$ being the height and width of the corresponding created image.

Moreover, the saliency map built from the eye-tracking recording can be considered as an image of one channel with a height $=45$ and width $=81$, the ratio between width and height being deduced from the VR headset resolution should be equal to $1.8$. One visual saliency map has been created for each trial with a corresponding EEG-Image. Moreover, the CNN being used to consider squared image, empty borders have been added at the top and the bottom of the visual saliency map. 

As explained earlier, the final framework is divided into two models: the generator and the discriminator. As in many other projects considering GAN, the training policy consist of competition between these two models. The aim of the discriminator being to distinguish the generated visual map among the ground truth and simultaneously to bias the discriminator by generating saliency map as close as possible to reality. The proposed architecture of the generator and discriminator is described in Table \ref{tab:ArchGen} and \ref{tab:ArchDis}.

Adam optimizers have been used to train the three models. A learning rate $=10^{-5}$ with momentum $=0$ have been considered to train both VAEs. For the GAN approach we consider two optimizers one for each part, a learning rate $=10^{-7}$ with a momentum $=10^{-5}$ for the generator and a learning rate $=10^{-5}$ with a momentum $=10^{-8}$ for the discriminator. The choice of a very low learning rate for the generator part has been motivated by the fact that the major part of the network was already trained and that we tried to not modify the model ability to extract feature from each modality (i.e. visual saliency heatmap and EEG images). The saliency VAE has been trained on $2000$ epochs and the EEG VAE on $3000$ epochs. The merged model has been trained on $1500$ epochs, however, the training was manually stopped if the convergence point was achieved. All the weights, codes and signals are freely available on \url{https://bit.ly/3pznZHm}.

All the models have been implemented with Pytorch library and were trained on one 24 GB Nvidia Titan RTX GPU. We consider a 5-fold cross-validation protocol evaluation on our dataset.

\begin{table}
    \begin{center}
        \begin{tabular}{l c}
        \multicolumn{2}{c}{\textbf{Discriminator}} \\
            \hline
            \hline
            Layer & Int/Out Channels\\
            \hline
            Conv2D-3 & 1/3\\
            Conv2D-3 & 3/32\\
            Conv2D-3 & 32/32\\
            Conv2D-3 & 32/32\\
            Maxpool-2 & 32/32\\
            Conv2D-3 & 32/64\\
            Conv2D-3 & 64/64\\
            Conv2D-3 & 64/64\\
            Maxpool-2 & 64/64\\
            Conv2D-3 & 64/128\\
            Conv2D-3 & 128/128\\
            Conv2D-3 & 128/128\\
            Maxpool-2 & 128/128\\
            Conv2D-3 & 128/256\\
            Conv2D-3 & 256/256\\
            Conv2D-3 & 256/256\\
            Flattening Layer & {}\\
            Linear & 256/64\\
            Linear & 64/1\\
            \hline
        \end{tabular}
    \end{center}
\caption{Architecture of the discriminator network.}
\label{tab:ArchDis}
\end{table}

\subsection{Metrics and evaluation}
Noting that the framework training is converging is one thing but evaluating the model's accuracy is another. After steps of hyper-parameters fine-tuning, it has been noted that the networks were making correct estimation either for training and testing sets. However, other metrics are required to have a fair comparison with existing methodology. As explained in the section \ref{sec:RW}, there are no works that use the considered modality, i.e. physiological measurement measurements, to estimate visual saliency map. For this reason, it was difficult to have a direct comparison with a well-made benchmark listing all the existing works in the state-of-the-art methods. However, estimating visual saliency map from images has been a major field of research over the last decade. Various metrics have also been defined to allow a fair comparison between the proposed models. Among, the metrics available in the MIT/Tuebingen Saliency Benchmark the following have been considered:
\begin{itemize}
    \item The AUC representing the area under the ROC curve. In the case of visual saliency estimation, the AUC has been adapted to suit with the problematic by considering a changing threshold for class estimation from a value between 0 and 1 (corresponding to saliency value). This adapted AUC is sometimes also called \textit{AUC-Judd} \cite{riche_saliency_2013}.
    \item The Normalized Scanpath Saliency (NSS) is a straightforward method to evaluate the model's ability to predict the visual attention map. It consists of the measurement of the distance between the normalized around 0 ground-truth saliency map and the model estimation \cite{peters_components_2005}.
    \item The binary cross-entropy (BCE) computing the distance between the prediction and the ground truth value in binary classification. Our problematic may be considered as a binary classification if we consider each pixel as a probability of being watched or not, this for this reason that we have also considered this metric.
    \item The Pearson's Correlation Coefficient (CC) is a linear correlation coefficient measuring the correlation between the ground truth and model estimation distributions \cite{bylinskii_what_2019}.
\end{itemize}

\section{Results and Discussion}

The proposed approach aiming to estimate visual attention from EEG has been assessed considering a quantitative and qualitative evaluation. Moreover, the effect of the discriminator has been investigated by considering the metrics with and without this model part.

\begin{table}
    \begin{center}
        \begin{tabular}{l l c c c c c}
            \hline
            Modality & Approach & AUC & NSS & CC \\
            \hline
            \hline
            \multirow{2}{*}{EEG}    & Our method (1) & 0.697 & 1.9869 & 0.383 \\
                                    & Our method (2) & 0.574 & 1.6891 & 0.251  \\
            \hline
            \multirow{3}{*}{Images} & UNISAL \cite{droste_unified_2020} & 0.877 & 2.3689 & 0.7851 \\
            &SalGAN \cite{pan_salgan_2018} & 0.8498 & 1.8620 & 0.6740 \\ 
            &SSR \cite{seo_nonparametric_2009} & 0.7064 & 0.9116 & 0.2999 \\
            \hline
        \end{tabular}
    \end{center}
\caption{Comparison of our approach results with the state of the art works for saliency map estimation. In the upper part of the table our approach with our dataset is presented and in the lower part different models results on MIT300 dataset \cite{judd_benchmark_2012} are shown. We presented in this table the results of two approaches: (1) the global model composed of Generator and Discriminator, (2) model composed only by the Generator.}
\label{tab:Res}
\end{table}

\subsection{Quantitative analysis}

The results for the saliency map estimation is presented in Table \ref{tab:Res}. As seen in this table, the different metrics presented in the previous section have been listed. However, the BCE is not mentioned in this table because it has been used to control the training, e.g. overfitting, training stuck in local minima, etc. 

As also seen in Table \ref{tab:Res}, our model presents encouraging results nevertheless it is not beating the state-of-the-art methods for visual saliency estimation from images. First, it is important to note that the considered modality is different from the one considered in this paper. Indeed, one of the main goals of this paper was to prove the existence of a relationship between electrophysiological signals and eye-tracking signals. Our results showing the generation of plausible heatmaps in several cases demonstrate this relationship. Moreover, it is interesting to note that our approach based on EEG presents similar results than older models estimating images, that may lead to better results in further works and/or with other signals.

\begin{figure}[b]
    \begin{center}
     \includegraphics[width=0.8\linewidth]{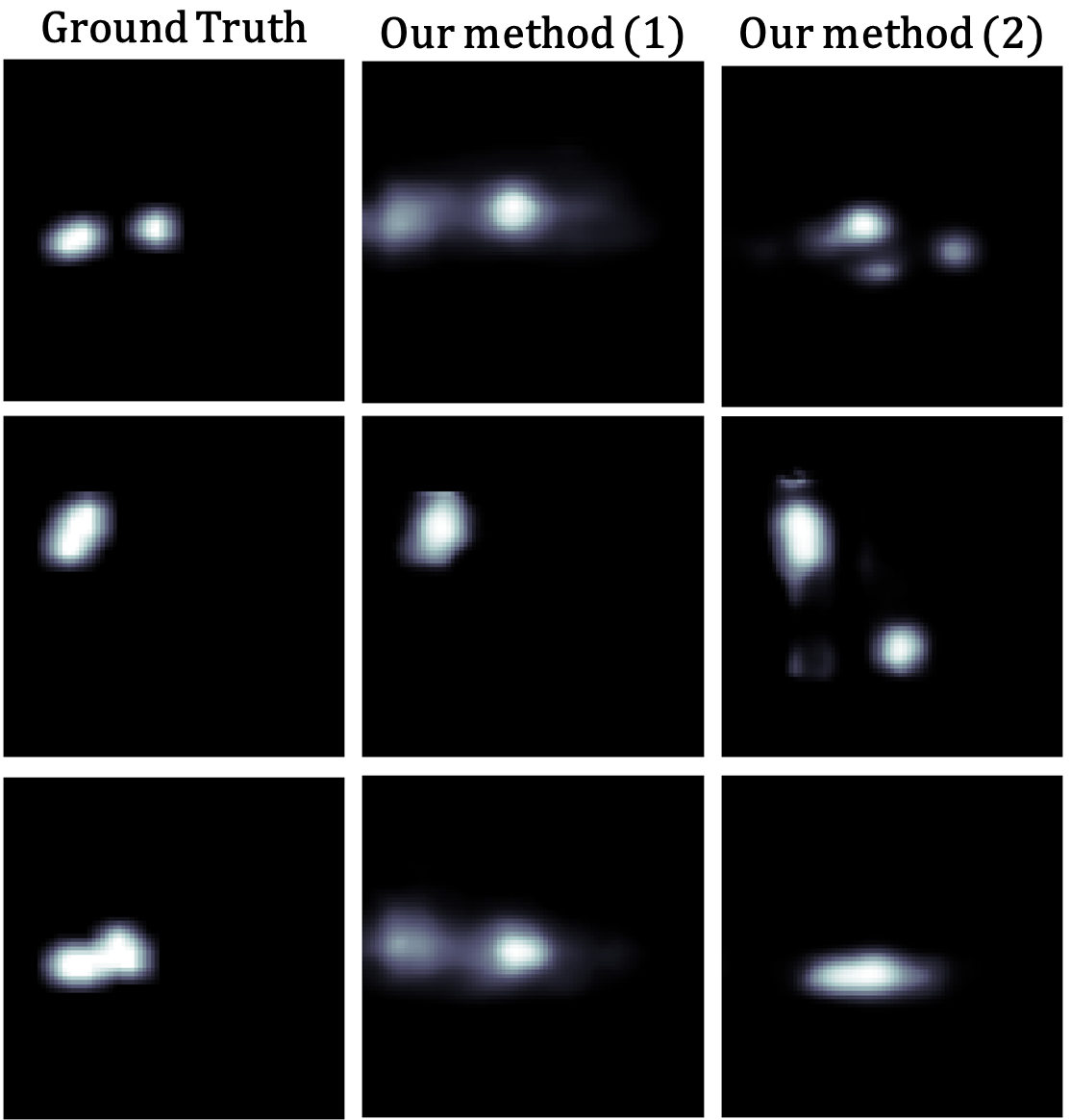}
    \end{center}
       \caption{Example of ground truth and estimated saliency map for the model trained with our dataset with and without considering the adversarial training. }
    \label{fig:ResQual}
\end{figure}

\subsection{Effect of the adversarial training}

In this subsection, an analysis of the improvements that the discriminator could bring will be made. For this purpose, two methodologies have been considered one with and one without taking into account the discriminator as it has already been considered in other studies \cite{pan_salgan_2018}.

As mentioned earlier, the goal of the discriminator is indirectly to force the generator to create images as similar as possible to the visual saliency map generated with eye-tracking signals. This phase is achieved through a competitive training process between the generator and the discriminator. In the Table \ref{tab:Res}, we note that the adversarial model, i.e. with the discriminator, present better results for the three metrics compared to the approach composed only from the generator. It seems that in addition to promote the generation of faithful maps, the adversarial could also help to make a better estimation.

\subsection{Qualitative analysis}

In parallel with the considered metrics considered for model evaluation, qualitative analysis can be made by visually comparing the saliency map estimated by the model and the ground truth heatmap as seen in Fig. \ref{fig:ResQual}. 

In addition to the improvements noticed in the metrics, a qualitative improvement for the generated images has also been observed with the adversarial training as shown in Fig. \ref{fig:ResQual}. The presence of discriminator helps the model to predict smoother saliency map and more real as shown in the second line of the Fig. \ref{fig:ResQual}. In addition to this improvement, this methodology seems also to help for detection of pattern locations as seen in the first line of the figure where the location of the centre of attention is closer from reality with the adversarial training compared to the other. 

\section{Conclusion}

In this paper, we presented a novel approach of visual attention estimation from electrophysiological recordings through their latent representation. To estimate the visual saliency map from EEG signals, we use a feature extraction in lower subspace and a representation under so-called EGG-images to feed a deep variational autoencoder model. The performance of our method has been evaluated on novel dataset acquired for physiological analysis purpose in VR from 32 participants during 15 minutes long session and, has been made publicly available. With the help of the proposed framework, the relationship between neurophysiological signal and eye-tracking has been proven. Further works will investigate the possible improvements that novel ML algorithms could bring.

In addition to demonstrating this relationship, this model could help for different applications in the field of attention and vigilance estimation, and could also be helpful for other estimation from EEG, e.g. emotion, motor imagery, etc.

\bibliographystyle{ieeetr}
\bibliography{Bibliography}

\end{document}